\pdfoutput=1

\documentclass[11pt]{article}

\usepackage[final]{coling}
\usepackage{pifont}
\usepackage{times}
\usepackage{latexsym}
\usepackage{multicol}
\usepackage{multirow}

\usepackage[T1]{fontenc}

\usepackage[utf8]{inputenc}

\usepackage{microtype}

\usepackage{inconsolata}

\usepackage{graphicx}
\usepackage{booktabs}
\usepackage{tikz}
\usepackage{xcolor}

\renewcommand{\arraystretch}{0.8} 
\usepackage[inline]{enumitem}
\usepackage{adjustbox}

\title{
Evaluating Pixel Language Models on Non-Standardized Languages
}

\usepackage{fontawesome5}
\newcommand{\coruna}{\faSun}
\newcommand{\lmu}{\faMountain}
\newcommand{\mcml}{\faRobot}

\author{Alberto Muñoz-Ortiz\kern1pt\textsuperscript{\coruna} \And
Verena Blaschke\kern1pt\textsuperscript{\lmu\kern1pt\mcml} \\
\textsuperscript{\coruna}\kern1ptUniversidade da Coruña, CITIC, Spain \\
\textsuperscript{\lmu}\kern1ptLMU Munich, Center for Information and Language Processing (CIS), Germany \\
\textsuperscript{\mcml}\kern1ptMunich Center for Machine Learning (MCML), Germany \\
\phantom{\textsuperscript{\mcml}}
\texttt{alberto.munoz.ortiz@udc.es, \{verena.blaschke,b.plank\}@lmu.de}
\And
Barbara Plank\kern1pt\textsuperscript{\lmu\kern1pt\mcml}}

\begin{document}
\maketitle

\begin{abstract}

We explore the potential of pixel-based models for transfer learning from standard languages to dialects. These models convert text into images that are divided into patches, enabling a continuous vocabulary representation that proves especially useful for out-of-vocabulary words common in dialectal data.
Using German as a case study, we compare the performance of pixel-based models to token-based models across various syntactic and semantic tasks. Our results show that pixel-based models outperform token-based models in part-of-speech tagging, dependency parsing and intent detection for zero-shot dialect evaluation by up to 26 percentage points in some scenarios, though not in Standard German. However, pixel-based models fall short in topic classification. These findings emphasize the potential of pixel-based models for handling dialectal data, though further research should be conducted to assess their effectiveness in various linguistic contexts.
\end{abstract}

\section{Introduction}
Despite being spoken by millions of people worldwide, dialects and other non-standard language forms are largely underrepresented in Natural Language Processing (NLP) systems.
Although pretrained language models (PLMs) achieve strong results for languages seen during training, where more data is available, their performance declines with out-of-domain dialects.

One of the primary factors contributing to the poor performance of PLMs on non-standard language varieties is tokenization,  as tokenizers frequently break dialects into sub-tokens that lack meaning. Modifying tokenization has been shown to improve performance on non-standard data \cite{aepli-sennrich-2022-improving, blaschke-etal-2023-manipulating, srivastava-chiang-2023-bertwich, srivastava-chiang-2023-fine}. 

In this context, dialectal variations can be viewed as a form of perturbation: tokenizing dialect data often produces tokens that are not meaningful. However, despite these variations, native speakers of the standard language can still comprehend dialects up to certain point due to linguistic and visual similarities. This suggests that visual cues may help models address the tokenization challenges posed by dialectal variations more effectively.

\begin{figure}[t]
\centering
\begin{tabular}{lcc}
a. & \textbf{Herzlich} & \textbf{willkommen!} \\
   & Herz\#\#lich  & willkommen, ! \\
& \multicolumn{2}{c}{\includegraphics[width=0.55\linewidth]{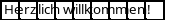}}
\\
& & \\
\\
b. & \textbf{Härzlech} & \textbf{wiukomme!} \\
   & Hä,\#\#rz,\#\#le,\#\#ch, & w\#\#iu\#\#komme,!
   \\
& \multicolumn{2}{c}{\includegraphics[width=0.55\linewidth]{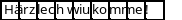}}
\end{tabular}

\caption{``Welcome!'' in Standard German (a) and the Swiss German Bern dialect (b) tokenized using DBMDZ German BERT and rendered and split in patches by PIXEL. Standard German is tokenized in a more meaningful way, whereas the Bernese dialect form results in multiple non-meaningful sub-tokens due to variations in spelling.}
\label{fig:tokenization}

\end{figure}

To address the limitations of traditional tokenization, visual text representations convert text into images divided into patches, offering an alternative approach.
Prior studies, such as \citet{salesky-etal-2021-robust} and \citet{rustlanguage}, have demonstrated that this approach effectively manages diverse scripts and languages without expanding the vocabulary, outperforming token-based approaches in syntactic tasks and machine translation, but not in semantic tasks. Strategies like structured rendering address this issue \cite{lotz-etal-2023-text}.

Following this approach, we explore the use of pixel-based models to enhance NLP performance on dialects. Using German as a case study, we pretrain a pixel-based model from scratch, which we release it publicly.\footnote{\url{https://huggingface.co/amunozo/pixel-base-german}}
We compare its performance to token-based models pretrained on the same data. Our results show that pixel-based models outperform token-based ones in syntactic tasks for zero-shot dialect evaluation but not in Standard German. In sentence-level semantic tasks, it excels in intent detection but is outperformed by token-based models in topic classification.

\section{Pixel-based models}
The \textbf{Pix}el-based \textbf{E}ncoder of \textbf{L}anguage \cite{rustlanguage}, or PIXEL, is a model that casts language modeling as a visual recognition task.
It is composed of:
\begin{enumerate*}[label=(\arabic*)]
    \item \textbf{Text Renderer}: The text rendered transforms a string of text into a RGB image, divided into equal-sized patches of 16x16 pixels.
    
    \item \textbf{Encoder}: Once the text is converted into patches, the image fed to a Vision Transformer \cite{he2022masked} architecture that processes it. PIXEL uses a 12-layer transformer with a total of 86M parameters as encoder.
    
    \item \textbf{Decoder or Task-Specific Head}: Instead of masking tokens, PIXEL masks spans of patches, using a decoder to reconstruct the masked patches as pretraining task. 
    This decoder is discarded after pretraining, being replaced by a task-specific classification head for fine-tuning.
\end{enumerate*}

Using a continuous vocabulary allows PIXEL to handle multiple languages and scripts without the need of expanding its vocabulary.
Also, this makes it robust to orthographic noise, such as typos or non-standard spellings, as it can generalize over orthographic variations which would break the tokenizations of token-based models. Finally, it avoids the high computational cost of a large vocabulary. 

However, despite comparable parameter counts, PIXEL requires more fine-tuning steps to converge than a token-based model for the same data \cite{rustlanguage}. 
Additionally, rendering text as images significantly increases disk space usage compared to plain text. While dynamic rendering during training or inference could alleviate storage concerns, it increases computational overhead.

\section{Experiments}
\subsection{Setup}
We investigate how pixel-based models pretrained on monolingual data compare to BERT \cite{devlin-etal-2019-bert} when evaluated on dialectal data. To this end, we pretrain a pixel-based model on monolingual German data from the DBMDZ corpus,\footnote{\url{https://huggingface.co/datasets/stefan-it/german-dbmdz-bert-corpus}} following \citet{rustlanguage}.\footnote{We use the code from \url{https://github.com/xplip/pixel}}

We choose a monolingual model as they perform competitively with multilingual models on dialect data \cite{bernier-colborne-etal-2022-transfer,castillo-lopez-etal-2023-analyzing}, and also due to computational constraints. We select German as our study language because of its wide range of dialectal variations, which show different degrees of standardization and are supported by available annotated data.

As a baseline, we use the cased\footnote{\url{https://huggingface.co/dbmdz/bert-base-german-cased}} and uncased\footnote{\url{https://huggingface.co/dbmdz/bert-base-german-uncased}} versions of DBMDZ German BERT, pretrained on the same data. For simplicity, we will refer to the models as \texttt{bert-cased}, \texttt{bert-uncased}, and \texttt{pixel}.
We fine-tune the three PLMs on part-of-speech (POS) tagging, dependency parsing, topic classification, and intent detection. For POS tagging, dependency parsing and topic classification, the models are trained on Standard German and evaluated on dialects. For intent detection, we train on both Standard German and dialects. The results were averaged over five runs. We followed]] the hyperparameters and setup for pretraining and fine-tuning from \citet{rustlanguage}. 
Detailed information about the datasets is available in the Appendix (Table~\ref{tab:datasets}).

\subsection{German non-standard varieties}
We evaluate our model on four non-standard language varieties related to German.
\textbf{Bavarian} and \textbf{Alemannic} are dialect groups spoken in the South of the German-speaking area. They are pronounced differently than Standard German (which is expressed when the dialects are written), and their vocabulary and grammar also show differences to Standard German \cite{merkle1993bairische, christen2019alemannisch}. Neither dialect group has any widely adopted orthography. 
For Alemannic, we focus on Swiss German and Alsatian German.
\textbf{Low Saxon} is a regional language spoken in Northern Germany and parts of the Netherlands. It is not standardized and encompasses multiple dialects \cite{wiesinger1983deutschedialekte}. Finally, we include \textbf{code-switched Turkish--German} data. The code-switching occurs on the level of morphemes, words and phrases.

\subsection{Syntactic tasks}
We cover POS tagging and dependency parsing together in this subsection, as both tasks are evaluated using the same datasets and show similar results.

\paragraph{Data}
For both POS tagging and dependency parsing, we use treebanks from Universal Dependencies (UD) \cite{nivre-etal-2020-universal, de-marneffe-etal-2021-universal} along with two non-UD datasets for Alemannic: NOAH's Corpus \cite{hollenstein-aepli-2014-compilation} and Alsatian Bisame GSW \cite{stih2020bisame}. The models were trained on two Standard German treebanks: GSD and HDT.

\paragraph{Results}

Table \ref{tab:syntactic_tasks_results} shows the POS-tagging accuracy and (un)labelled attachment scores (UAS, LAS) of the models trained on Standard German and evaluated on dialects.

\begin{table}
\centering
\adjustbox{max width=\linewidth}{%
\begin{tabular}{@{}llrrrrrr@{}}
\toprule
 & & \multicolumn{3}{c}{\textbf{GSD}} & \multicolumn{3}{c}{\textbf{HDT}} \\
\cmidrule(lr){3-5} \cmidrule(lr){6-8}
\textbf{Language} & \textbf{Model} & \textbf{Acc}. & \textbf{UAS} & \textbf{LAS} & \textbf{Acc}. & \textbf{UAS} & \textbf{LAS} \\
\midrule
\multirow{3}{*}{German GSD} & \texttt{bert-cased} & \textbf{96.2}& 89.6 & 85.6 & 90.5 & 83.8 & 77.9 \\
 & \texttt{bert-uncased} & 96.2 & \textbf{89.8} & \textbf{85.8} & 90.9 &  \textbf{84.3} & \textbf{78.5} \\
 & \texttt{pixel} & 95.2 & 86.1 & 81.3 & \textbf{91.5} & 82.5 & 76.3 \\
\midrule
\multirow{3}{*}{German HDT} & \texttt{bert-cased} & \textbf{89.9} & \textbf{89.5}  & \textbf{84.1} & \textbf{98.6} & 97.8  & \textbf{96.9 }\\
 & \texttt{bert-uncased} & 89.8 & 89.3 & 83.9 & 98.5 & \textbf{97.8} & 96.9 \\
 & \texttt{pixel} & 89.6 & 88.5  & 82.6 &  98.5 & 96.9 & 95.8 \\
\midrule
\midrule
\multirow{3}{*}{\begin{tabular}[c]{@{}l@{}}Bavarian\\MaiBaam\end{tabular}} & \texttt{bert-cased} & \textbf{54.6} &  53.0 & 35.6 &  43.1 & 32.6 & 23.2 \\
   & \texttt{bert-uncased} &  46.1 & 44.7 & 28.7 & 33.1 &  26.2 & 18.4 \\
   & \texttt{pixel} &  54.5 & \textbf{54.0} & \textbf{38.3} & \textbf{48.4} & \textbf{39.5} & \textbf{29.4} \\
\midrule
\multirow{3}{*}{\begin{tabular}[c]{@{}l@{}}Low Saxon\\LSDC\end{tabular}} & \texttt{bert-cased} &  33.3 & \textbf{34.1} & 17.8 & 17.3 &  9.5 & 5.9 \\
      & \texttt{bert-uncased} &  33.6 & 32.7 & 16.8 &  17.4 & 8.6 & 5.3 \\
      & \texttt{pixel} &  \textbf{37.2} & 32.9 & \textbf{18.1} &  \textbf{23.9} & \textbf{14.1} & \textbf{8.2} \\
\midrule
\multirow{3}{*}{\begin{tabular}[c]{@{}l@{}}Turkish--German\\SAGT\end{tabular}} & \texttt{bert-cased} &  \textbf{56.1} & \textbf{42.5} & \textbf{32.4} &  54.7 & \textbf{38.6} & \textbf{31.8} \\
           & \texttt{bert-uncased} &  54.4 & 40.8 & 32.1 & 53.8 &  38.3 & 31.5 \\
           & \texttt{pixel} & 55.6 &  40.2 & 29.8 & \textbf{55.5} &  37.3 & 29.9 \\
\midrule
\multirow{3}{*}{\begin{tabular}[c]{@{}l@{}}Swiss German\\UZH\end{tabular}} & \texttt{bert-cased} &  58.2 & 50.1 & 33.3 &  45.1 & 31.3 & 22.6 \\
        & \texttt{bert-uncased} & 50.6 &  41.9 & 26.6 & 35.5 & 26.7 & 18.4 \\
        & \texttt{pixel} & \textbf{59.2} & \textbf{51.5} & \textbf{35.8} &  \textbf{54.9} & \textbf{39.6} & \textbf{29.7} \\
\midrule
\multirow{3}{*}{\begin{tabular}[c]{@{}l@{}}Swiss German\\NOAH's\end{tabular}} & \texttt{bert-cased} &  63.1 & --- & --- & 54.2 & --- & --- \\
        & \texttt{bert-uncased} & 55.3 & --- & --- & 45.2 & --- & --- \\
        & \texttt{pixel} &  \textbf{63.4} & --- & --- & \textbf{62.1} & --- & --- \\
\midrule
\multirow{3}{*}{\begin{tabular}[c]{@{}l@{}}Alsatian\\BISAME\end{tabular}} & \texttt{bert-cased} &  45.8 & --- & --- & 34.1 & --- & --- \\
        & \texttt{bert-uncased} & 49.6 & --- & --- & 30.4 & --- & --- \\
        & \texttt{pixel} &  \textbf{53.3} & --- & --- & \textbf{48.2} & --- & --- \\
\bottomrule
\end{tabular}
}
\caption{\textbf{POS tagging and dependency parsing performance} (in~\%) of models trained on German GSD and HDT and tested on different dialects.}
\label{tab:syntactic_tasks_results}
\end{table}

When trained on the GSD dataset, \texttt{bert-cased} performs best on Standard German treebanks and Turkish German code-switching, while \texttt{pixel} outperforms BERT on the Alemannic treebanks, Low Saxon, and Alsatian. It also performs comparably to BERT on Bavarian for POS tagging and outperforms it for dependency parsing.

When trained on HDT, \texttt{pixel} widens the performance gap, outperforming BERT on most dialect treebanks except HDT itself. Although all models experience a decline in accuracy, UAS and LAS, \texttt{pixel} demonstrates greater robustness.

In both tagging and parsing, \texttt{pixel} outperforms BERT during zero-shot evaluation on German dialects. 
BERT shows contrasting results: \texttt{bert-uncased} achieves the best performance on Standard German but performs significantly worse on dialects. Since nouns in German are capitalized, \texttt{bert-cased} likely leverages this feature to compensate for poor tokenization when processing dialects.

\paragraph{Accuracy per POS tag}
To gain deeper insights, we calculate the average accuracy per POS tag for each model trained on the Standard German treebanks and evaluated on dialects (Table \ref{tab:average_accuracy_tags}).

For GSD, \texttt{pixel} outperforms BERT for all tags except \texttt{DET}, \texttt{NOUN}, \texttt{PROPN}, \texttt{ADV}, and \texttt{X}, which is the only tag where \texttt{bert-cased} outperforms \texttt{pixel} when trained on HDT. While these results are difficult to fully explain, there are plausible explanations for certain tags. For example, memorization plays a role for \texttt{PROPN}, which favors token-based models, and proper nouns might vary less between languages. Furthermore, words tagged as \texttt{NUM} and \texttt{PUNCT} exhibit visual similarities within each group, which benefits \texttt{pixel}.

\newcommand{\postag}[1]{\texttt{\small{#1}}}
\begin{table*}
\centering
\adjustbox{max width=\linewidth}{%
\begin{tabular}{@{}llrrrrrrrrrrrrrrrrr@{}}
\toprule
\textbf{Src} & \textbf{Model} & \postag{ADJ} & \postag{ADP} & \postag{ADV} & \postag{AUX} & \postag{CCON\rlap{J}} & \postag{DET} & \postag{INTJ} & \postag{NOUN} & \postag{NUM} & \postag{PART} & \postag{PRON} & \postag{PROPN} & \postag{PUNCT} & \postag{SCONJ} & \postag{SYM} & \postag{VERB} & \postag{X} \\
\midrule
\multirow{3}{*}{GSD} & {\texttt{bert-cased}}   &  42.9 &  55.5 &  \textbf{47.1} &  29.8 &  69.6 &  \textbf{31.2} &   0.0 &  \textbf{51.5} &  58.7 &  16.7 &  40.4 &  \textbf{89.4} &  99.3 &  41.0 &  5.6 &  43.8 &  \textbf{11.2} \\
& \texttt{bert-uncased} &  32.5 &  55.1 &  44.1 &  22.9 &  71.2 &  27.6 &   0.0 &  40.9 &  58.2 &  18.4 &  41.1 &  87.2 &  99.3 &  \textbf{43.5} &  0.3 &  39.6 &   8.0 \\
& \texttt{pixel}        &  \textbf{49.3} &  \textbf{60.8} &  45.2 &  \textbf{31.7} &  \textbf{75.9} &  26.5 &   0.0 &  50.7 &  \textbf{63.9} &  \textbf{21.8} &  \textbf{46.5} &  87.1 &  \textbf{99.8} &  40.4 &  0.0 &  \textbf{52.8} &   5.1 \\
\midrule
\multirow{3}{*}{HDT} & \texttt{\texttt{bert-cased}}   &  39.5 &  31.8 &  21.8 &  19.6 &  51.7 &  17.9 &  14.9 &  49.9 &  59.8 &  13.3 &  33.3 &  52.3 &  97.1 &  29.6 &  0.0 &  29.4 &  \textbf{66.6} \\
& \texttt{bert-uncased} &  28.9 &  30.5 &  21.1 &  18.8 &  50.5 &  15.6 &  17.6 &  31.6 &  59.5 &  13.0 &  26.2 &  53.5 &  96.3 &  27.2 &  0.0 &  20.7 &  58.8 \\
& \texttt{pixel}        &  \textbf{50.3} &  \textbf{47.7} &  \textbf{32.3} &  \textbf{28.8} &  \textbf{56.6} &  \textbf{21.9} &  \textbf{21.9} &  \textbf{53.3} &  \textbf{64.8} &  \textbf{17.5} &  \textbf{36.9} &  \textbf{63.5} &  \textbf{99.4} &  \textbf{35.1} &  0.0 &  \textbf{45.3} & 64.2 \\
\bottomrule
\end{tabular}
}
\caption{\textbf{Average accuracy per POS tag} (in~\%) for models trained on GSD and HDT when evaluating on dialect treebanks.}
\label{tab:average_accuracy_tags}
\end{table*}

\begin{figure}
    \centering
    \includegraphics[width=\linewidth]{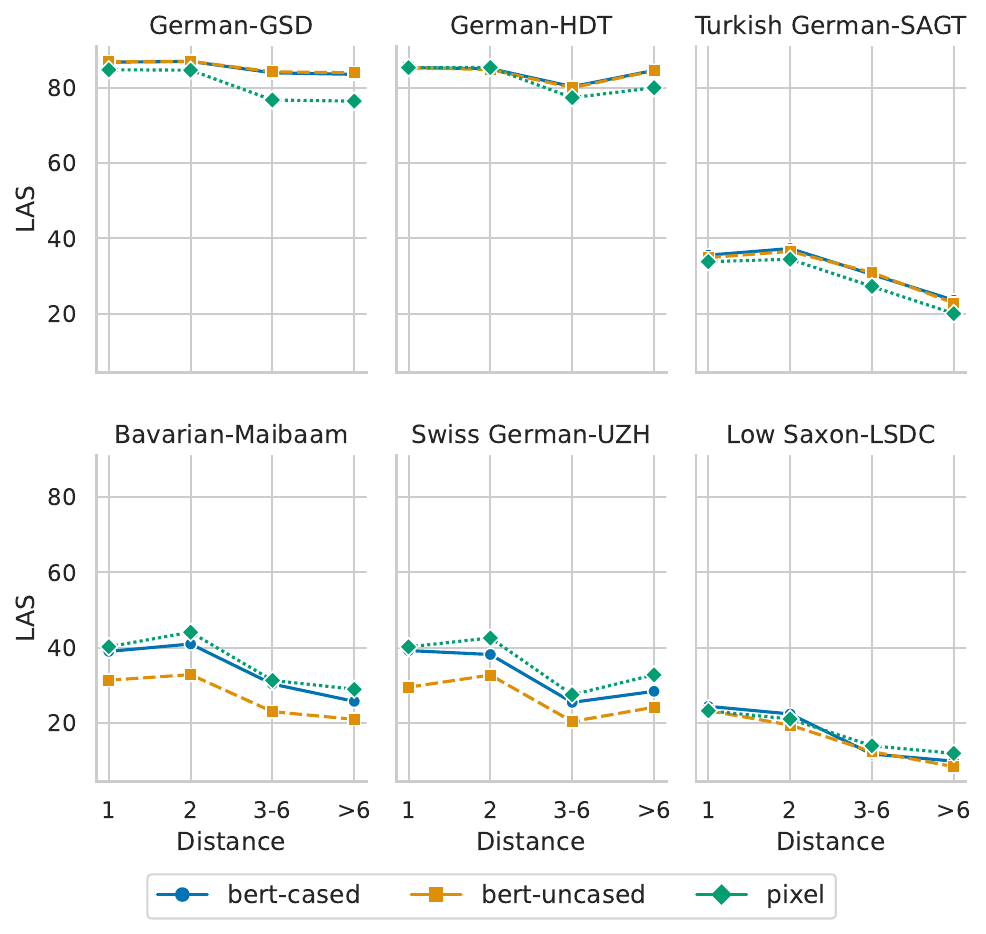}
    \caption{\textbf{Labelled attachment scores (in~\%) for different dependency distances} for models trained on German GSD.}
    \label{fig:dep_per_distance_GSD}
\end{figure}

\paragraph{LAS per dependency length}
To help explain why relative performance in POS tagging is better than in dependency parsing, we measured LAS performance based on dependency lengths, as in \citet{rustlanguage}. Figure \ref{fig:dep_per_distance_GSD} plots LAS per dependency length for models trained on GSD. Results on Standard German and Turkish German diverge as dependency length grows. Moreover, \texttt{bert-cased} and \texttt{bert-uncased} show similar results.

For dialects, however, \texttt{pixel} achieves higher LAS across all lengths for Bavarian and Alemannic, and the performance gap neither consistently widens nor narrows. For Low Saxon, where overall results are lower, \texttt{pixel}'s performance relative to BERT improves with increasing distance, surpassing BERT at distances of 3 and beyond, but not at shorter distances.
Interestingly, the \texttt{pixel}'s poorer handling of long dependencies observed for Standard German and in \citet{rustlanguage} is not observed when evaluation on dialects.

Lastly, we observe that \texttt{bert-uncased} performs considerably worse than \texttt{bert-cased}, unlike for Standard German. 

\subsection{Topic Classification}
\paragraph{Data} We use SwissDial \cite{dogan2021swissdial}, a parallel corpus of Standard German and 8 Swiss German dialects, annotated with topic labels. We trained the models on four different datasets: \texttt{all} (a combination of Standard German and the 8 Swiss German dialects), \texttt{ch} (only the 8 Swiss German dialects), \texttt{de} (only Standard German), and \texttt{gr} (the Grisons dialect, which is the dialect that contains the most examples in the dataset and has the largest number of them in common with the Standard German data). We evaluate all the models on each variety.

\paragraph{Results} The results are shown in Table \ref{tab:model_performance_swissdial}. BERT models outperform \texttt{pixel} in most setups: \texttt{bert-cased} performs best when trained only on Standard German, and \texttt{bert-uncased} in the rest.

\texttt{Pixel} only outperforms BERT in two cases: on the Basel-Stadt dialect when trained on \texttt{all}, and on the Grisons dialect when trained on \texttt{de}. Although two improvements are not enough to draw conclusions, we observe that in both cases, the models have been trained on at least some Standard German data and evaluated on Swiss German dialects. 

Transfer learning from Standard German to Swiss dialects is competitive with BERT, but the opposite is not true: BERT models trained on dialect data and tested on Standard German outperform \texttt{pixel}, likely due to more efficient tokenization.
\begin{table}
\centering
\adjustbox{max width=\linewidth}{%
\begin{tabular}{@{}llrrrrrrrrr@{}}
\toprule
 \textbf{Src} & \textbf{Model} & \textbf{de} & \textbf{ag} & \textbf{be} & \textbf{bs} & \textbf{gr} & \textbf{lu} & \textbf{sg} & \textbf{vs} & \textbf{zh} \\
\midrule
\multirow{3}{*}{\texttt{all}} 
    & \texttt{bert-cased}   & 50.5 & 59.4 & 58.1 & 61.0 & 46.1 & 60.4 & 60.6 & 58.8 & 55.1 \\
    & \texttt{bert-uncased} & \textbf{50.7} & \textbf{63.5} & \textbf{58.1} & 60.5 & \textbf{46.7} & \textbf{61.9} & \textbf{62.2} & \textbf{60.9} & \textbf{58.1} \\
    & \texttt{pixel}        & 44.5 & 57.6 & 57.1 & \textbf{60.3} & 43.8 & 57.3 & 58.6 & 58.1 & 55.1 \\
\midrule
\multirow{3}{*}{\texttt{ch}} 
    & \texttt{bert-cased}   & 42.2 & 55.6 & 58.6 & \textbf{63.9} & 47.8 & 58.6 & 61.4 & 59.6 & 56.9 \\
    & \texttt{bert-uncased} & \textbf{45.7} & \textbf{59.6} & 58.4 & 61.8 & \textbf{48.2} & \textbf{62.7} & \textbf{62.2} & \textbf{62.6} & \textbf{60.5} \\
    & \texttt{pixel}        & 36.3 & 57.1 & \textbf{58.9} & 58.2 & 41.1 & 56.8 & 57.8 & 58.6 & 54.9 \\
\midrule
\multirow{3}{*}{\texttt{de}} 
    & \texttt{bert-cased}   & 50.0 & \textbf{34.0} & \textbf{34.2} & \textbf{40.0} & 30.4 & \textbf{30.3} & \textbf{41.7} & \textbf{32.8} & \textbf{32.0} \\
    & \texttt{bert-uncased} & \textbf{52.4} & 22.3 & 19.5 & 25.6 & 34.4 & 21.1 & 29.2 & 26.5 & 25.2 \\
    & \texttt{pixel}        & 45.4 & 30.9 & 30.1 & 36.2 & \textbf{35.3} & 28.0 & 33.8 & 32.1 & 30.1 \\
\midrule
\multirow{3}{*}{\texttt{gr}} 
    & \texttt{bert-cased}   & 44.0 & 46.7 & 49.1 & 49.7 & 48.2 & \textbf{48.6} & 51.2 & 49.5 & 45.6 \\
    & \texttt{bert-uncased} & \textbf{47.3} & \textbf{50.0} & \textbf{51.7} & \textbf{50.8} & \textbf{49.1} & 44.2 & \textbf{52.2} & \textbf{50.8} & \textbf{45.7} \\
    & \texttt{pixel}        & 41.0 & 42.1 & 48.8 & 44.4 & 43.8 & 44.2 & 48.1 & 41.9 & 37.5 \\
\bottomrule
\end{tabular}
}
\caption{\textbf{Topic classification accuracy} (in~\%) for models trained in the four training setups and evaluated on various targets in the SwissDial dataset. Key: de:~Standard German, ag:~Aargau, be:~Bern, bs:~Basel-Stadt, gr:~Grisons, lu:~Lucerne, sg:~St. Gallen, vs:~Valais, zh:~Zurich, ch:~all Swiss dialects.}
\label{tab:model_performance_swissdial}
\end{table}

\subsection{Intent Detection}
\paragraph{Data} We use xSID 0.5 \cite{van-der-goot-etal-2021-masked, aepli-etal-2023-findings, winkler-etal-2024-slot}, a cross-lingual slot and intent detection dataset. We use the machine-translated German training set, the (human-translated) German test set (de) in addition to one Swiss German (gsw) and two Bavarian (de\_ba, de\_st) test sets. 
We additionally use the translated and naturalistic Bavarian intent classification test sets introduced by \citet{winkler-etal-2024-slot} (MAS:de-ba, nat:de-ba).

\paragraph{Results}
Table \ref{tab:indent_detection} shows the accuracies on the test sets. 
\texttt{Pixel} outperforms BERT for every dialect except MAS:de-ba. 
The differences are substantial for Swiss German, and notable but more modest for the Bavarian dialects. These results show promising signs of \texttt{pixel} for certain semantic tasks when evaluating on dialects.

\begin{table}
\centering
\adjustbox{max width=\linewidth}{%
\begin{tabular}{@{}lrrrrrr@{}}
\toprule
 \textbf{Model}& \textbf{de} & \textbf{M:ba} & \textbf{nat:ba} & \textbf{de-ba} & \textbf{de-st} & \textbf{~gsw} \\
\midrule
     \texttt{bert-cased}   & 98.2 & 20.0 & 65.4 & 78.6 & 79.8 & 62.4 \\
     \texttt{bert-uncased} & \textbf{99.2} & \textbf{26.2} & 65.4 & 74.8 & 76.2 & 57.6 \\
     \texttt{pixel}        & 97.0 & 23.3 & \textbf{76.2} & \textbf{79.2} & \textbf{88.6} & \textbf{83.8} \\
\bottomrule
\end{tabular}
}
\caption{\textbf{Intent classification accuracy} (in~\%) for models trained on Standard German and evaluated on German dialects.}
\label{tab:indent_detection}
\end{table}

\subsection{Discussion}
{Pixel-based models has been show to underperform in sequence classification tasks \cite{rustlanguage}. This can be attributed to the uniform rendering of text into 16$\times$16 pixel patches. Unlike token classification, where words consistently map to similar patches, sequence classification introduces variability due to sentence progression, leading to slight variations in representation for the same word. \citet{lotz-etal-2023-text} explored patch multiplicity reduction strategies, like pairing two characters per patch or using monospace fonts.

{In our experiments, results in topic classification match this trend. However, pixel-based models outperform token-based ones in intent detection. This disparity may arise due to the dataset complexity, as topic classification on Standard German ($\sim$50\% accuracy) is inherently more challenging than intent detection ($\sim$100\% accuracy).

\section{Conclusion}
We presented a study on the use of pixel-based pretrained language models for zero-shot dialect evaluation, using German as a case study. Pixel-based models achieved higher scores than both cased and uncased token-based models when trained on Standard German and evaluated on German dialects for POS tagging, dependency parsing, and intent detection. However, they lagged behind token-based models in topic classification and for all tasks when evaluated on Standard German.

Pixel-based models showed promising results, particularly in intent detection, highlighting their potential in handling linguistic diversity. While their performance in topic classification indicates areas for further refinement and study, these models offer a novel approach to addressing dialectal NLP tasks. The current limitations in the availability of dialectal datasets present challenges for conducting a comprehensive evaluation, but we argue that pixel models have the potential to expand their utility across a broader range of NLP tasks, providing robust and adaptable language processing solutions, especially in low-resource contexts.

This work highlights the potential of pixel-based models in tackling the challenges posed by non-standard language varieties. With sufficient computational resources and data, multilingual pixel-based models could prove valuable for low-resource languages by bypassing tokenization and vocabulary limitations. However, the resource constraints commonly associated with low-resource language research \cite{ahia-etal-2021-low-resource} may hinder their practical adoption. While the model's success with German dialects is encouraging, its generalizability to other languages and dialects remains uncertain.

\section*{Limitations}
This study is constrained by several factors. Due to the high computational cost of pretraining language models, we focused on a single language, German, which limited our ability to explore other languages with different morphological or syntactic structures, or multilingual approaches. The scarcity of annotated data for dialectal varieties further restricted the scope of our experiments, excluding potential tasks and languages. While we included multiple dialects, the data imbalance and annotations quality may have introduced biases.

While our results show promise for German dialects, the generalization of these findings to other languages and language families remains uncertain and requires further investigation.

\section*{Acknowledgments}
We thank Huangyan Shan for her assistance in identifying and cleaning appropriate datasets for this work, as well as for contributing to some preliminary visualizations.

This work was funded by the European Research Council (ERC)  Consolidator Grant DIALECT 101043235; SCANNER-UDC (PID2020-113230RB-C21) funded by MICIU/AEI/10.13039/501100011033; Xunta de Galicia (ED431C 2024/02); GAP (PID2022-139308OA-I00) funded by MICIU/AEI/10.13039/501100011033/ and by ERDF, EU; Grant PRE2021-097001 funded by MICIU/AEI/10.13039/501100011033 and by ESF+ (predoctoral training grant associated to project PID2020-113230RB-C21); LATCHING (PID2023-147129OB-C21) funded by MICIU/AEI/10.13039/501100011033 and ERDF; and Centro de Investigación de Galicia ‘‘CITIC’’, funded by the Xunta de Galicia through the collaboration agreement between the Consellería de Cultura, Educación, Formación Profesional e Universidades and the Galician universities for the reinforcement of the research centres of the Galician University System (CIGUS).

\bibliography{custom, anthology}

\appendix

\section{Datasets}
\label{sec:appendix-datasets}
Table~\ref{tab:datasets} lists the downstream task datasets used for training and evaluation.

\begin{table*}
\centering
\renewcommand{\arraystretch}{1.3}
\adjustbox{max width=\textwidth}{%
\begin{tabular}{@{}ll@{\hspace{-5pt}}rr@{\hspace{8pt}}rr@{\hspace{8pt}}rr@{\hspace{8pt}}ll@{}}
\toprule
 &  & \multicolumn{2}{c}{\textbf{Train}} & \multicolumn{2}{c}{\textbf{Dev}} & \multicolumn{2}{c}{\textbf{Test}} &  &  \\
\cmidrule(lr){3-4} \cmidrule(lr){5-6} \cmidrule(lr){7-8}
\textbf{Dataset} & \textbf{Varieties} & \textbf{Sent} & \textbf{Word} & \textbf{Sent} & \textbf{Word} & \textbf{Sent} & \textbf{Word} & \textbf{Task} & \textbf{Licence} \\
\midrule
\multicolumn{10}{c}{\textit{Token-level tasks}} \\
UD German HDT 2.10 & German & 153.0k & 2799.1k & 18.4k & 324.8k & 18.5k & 331.7k & P, D & \href{https://github.com/UniversalDependencies/UD_German-HDT/blob/master/LICENSE.txt}{BY-SA 4.0}\\[-6pt]
\cite{borges-volker-etal-2019-hdt} &&&&&&&&& (annotations)\\
UD German GSD 2.10 & German & 13.8k & 263.8k & 799 & 12.5k & 977 & 16.5k & P, D & \href{https://github.com/UniversalDependencies/UD_German-GSD/blob/master/LICENSE.txt}{BY-SA 4.0} \\[-6pt]
\cite{mcdonald-etal-2013-universal}\\
UD Swiss German UZH 2.10 & Swiss German & ---\phantom{*} & ---\phantom{*} & ---\phantom{*} & ---\phantom{*} & 100 & 1.4k & P, D & \href{https://github.com/UniversalDependencies/UD_Swiss_German-UZH/blob/master/LICENSE.txt}{BY-SA 4.0} \\[-6pt]
\cite{aepli2018parsing}\\
UD Low Saxon LSDC 2.10 & Low Saxon & ---\phantom{*} & ---\phantom{*} & ---\phantom{*} & ---\phantom{*} & 83 & 2.5k & P, D & \href{https://github.com/UniversalDependencies/UD_Swiss_German-UZH/blob/master/LICENSE.txt}{BY-SA 4.0} \\[-6pt]
\cite{siewert-rueter-2024-low} \\
UD Turkish German SAGT 2.10 & Code-switched & ---* & ---* & ---* & ---* & 805 & 13.9k & P, D & \href{https://github.com/UniversalDependencies/UD_Turkish_German-SAGT/blob/master/LICENSE.txt}{BY-SA 4.0} \\[-6pt]
\cite{cetinoglu2022} & Turkish--German\\
UD Bavarian MaiBaam 2.14 & Bavarian & ---\phantom{*} & ---\phantom{*} & ---\phantom{*} & ---\phantom{*} & 1.1k & 15.0k & P, D & \href{https://github.com/UniversalDependencies/UD_Bavarian-MaiBaam/blob/master/LICENSE.txt}{BY-SA 4.0} \\[-6pt]
\cite{blaschke-etal-2024-maibaam}\\
NOAH's corpus & Swiss German & ---\phantom{*} & ---\phantom{*} & ---\phantom{*} & ---\phantom{*} & 7.3k & 113.6k & P & \href{https://github.com/noe-eva/NOAH-Corpus/blob/master/LICENSE}{BY 4.0}\\[-6pt]
\cite{hollenstein-aepli-2014-compilation} &&&&&&&&& (annotations)\\
Alsatian Bisame GSW & Alsatian & ---\phantom{*} & ---\phantom{*} & ---\phantom{*} & ---\phantom{*} & 382 & 8.2k & P & \href{https://creativecommons.org/licenses/by-nc-sa/3.0/fr/deed.en}{BY-NC-SA 3.0} \\[-6pt]
\cite{stih2020bisame}\\
\midrule
\multicolumn{10}{c}{\textit{Sentence-level tasks}} \\
SwissDial & German,  & \multicolumn{6}{c}{\textit{2.5k--4.1k sents per variety, split 80:10:10}} & T & \href{https://creativecommons.org/licenses/by-nc/4.0/deed.en}{BY-NC 4.0} \\[-6pt]
\cite{dogan2021swissdial} & 8$\times$Swiss German & \\
xSID 0.5  & German, 2$\times$Bavarian, & ---\phantom{*} &  & ---* & & 4$\times$500 & & I & \href{https://github.com/mainlp/xsid/blob/main/LICENSE}{BY 4.0}\\[-6pt]
(\citealp{van-der-goot-etal-2021-masked}; &  Swiss German \\[-6pt]
\multicolumn{2}{l}{\citealp{aepli-etal-2023-findings, winkler-etal-2024-slot})}  \\
NaLiBaSID MAS:de-ba & Bavarian & ---\phantom{*} &  & ---\phantom{*} &  & 2.0k & & I & not specified \\[-6pt]
\cite{winkler-etal-2024-slot}\\
NaLiBaSID nat:de-ba & Bavarian & ---\phantom{*} &  & ---\phantom{*} & & 315 & & I & not specified \\[-6pt]
\cite{winkler-etal-2024-slot}\\
\bottomrule
\end{tabular}
}
\caption{\textbf{Training and evaluation datasets used in our experiments.}
P\,=\,part-of-speech tagging, D\,=\,dependency parsing,
T\,=\,topic classification,
I\,=\,intent classification.
*The original dataset comes with training and/or development splits, but we do not use them.}
\label{tab:datasets}
\end{table*}

\end{document}